\documentclass[wcp]{jmlr}

\usepackage{longtable}% for long tables

\usepackage{booktabs}
\usepackage{multirow}
\usepackage{colortbl}
\usepackage{pifont}
\usepackage{caption}
\def\bw{\mathbf{w}}

\usepackage{lineno}
\makeatletter
\let\Ginclude@graphics\@org@Ginclude@graphics 
\makeatother

\jmlrvolume{}
\jmlryear{2024}
\jmlrworkshop{ACML 2024}

\title[FedLF: Federated Long-Tailed Learning]{FedLF: Adaptive Logit Adjustment and Feature Optimization in Federated Long-Tailed Learning}
 \author{
 \Name{Xiuhua Lu} \Email{xhlu@qfnu.edu.cn} \and
 \Name{Peng Li} \Email{lipeng@qfnu.edu.cn}\thanks{Xiuhua Lu and Peng Li share equal contributions.}\thanks{Corresponding Authors: Peng Li and Xuefeng Jiang} \\
 \addr School of Cyber Science and Engineering, Qufu Normal University
 \AND
\Name{Xuefeng Jiang} \Email{jiangxuefeng21b@ict.ac.cn}\\
\addr Institute of Computing Technology, Chinese Academy of Sciences \& University of Chinese Academy of Sciences
}
\editors{}

\begin{document}

\maketitle

\begin{abstract}
Federated learning offers a paradigm to the challenge of preserving privacy in distributed machine learning. However, datasets distributed across each client in the real world are inevitably heterogeneous, and if the datasets can be globally aggregated, they tend to be long-tailed distributed, which greatly affects the performance of the model. The traditional approach to federated learning primarily addresses the heterogeneity of data among clients, yet it fails to address the phenomenon of class-wise bias in global long-tailed data. This results in the trained model focusing on the head classes while neglecting the equally important tail classes. Consequently, it is essential to develop a methodology that considers classes holistically. To address the above problems, we propose a new method FedLF, which introduces three modifications in the local training phase: adaptive logit adjustment, continuous class centred optimization, and feature decorrelation. We compare seven state-of-the-art methods with varying degrees of data heterogeneity and long-tailed distribution. Extensive experiments on benchmark datasets CIFAR-10-LT and CIFAR-100-LT demonstrate that our approach effectively mitigates the problem of model performance degradation due to data heterogeneity and long-tailed distribution. our code is available at \href{https://github.com/18sym/FedLF}{https://github.com/18sym/FedLF}.
\end{abstract}
\begin{keywords}
Federated learning, long-tailed distribution, data heterogeneity
\end{keywords}

\section{Introduction}

Due to the availability of large-scale data ~\cite{DBLP:conf/cvpr/DengDSLL009}, ~\cite{DBLP:conf/eccv/LinMBHPRDZ14}, ~\cite{DBLP:conf/cvpr/HornASCSSAPB18}and privacy-preserving policies~\cite{DBLP:conf/sp/MohasselZ17}, there is no way for data distributed across clients to be sent to a central server for model training. To address this problem, federated learning (FL) enables multiple clients to collaboratively train a global model without uploading their local private data to the server. As deep learning continues to evolve, FL shows great potentials as a privacy-preserving and communication-efficient framework in various application domains ~\cite{DBLP:journals/twc/ZengSCSB22}, \cite{diagnostics13091532}.

\begin{figure}[!t]
	\centering
	\includegraphics[width=.8\textwidth, trim=0 50 0 0, clip]{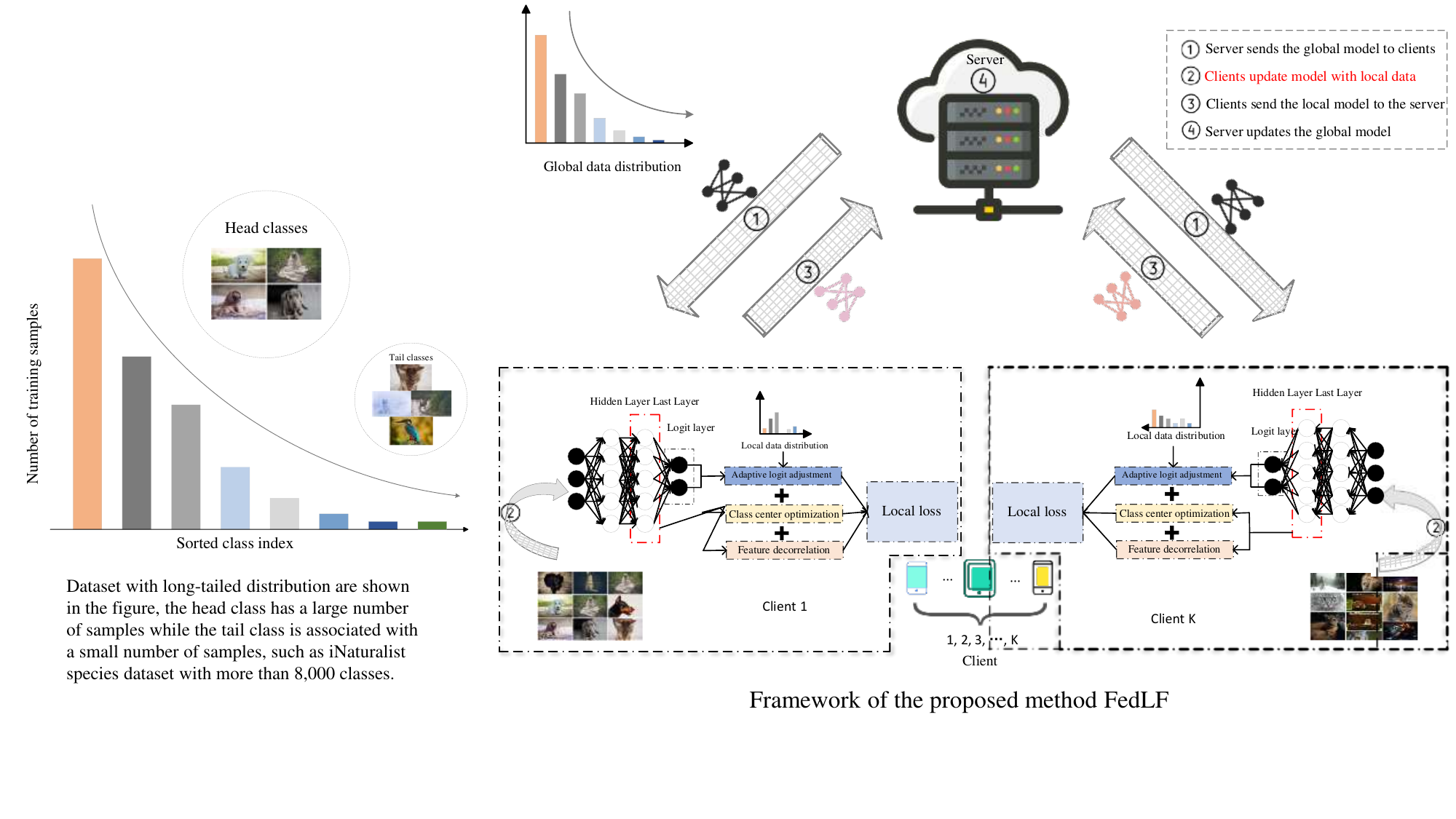}
	\caption{Left: global long-tailed distribution; Right: framework of FedLF}
	\label{long-tailed and framework}	
\end{figure}

However, in FL, the data sets of the participating clients come from different sources, and data heterogeneity is inevitable ~\cite{fedcrac,fedtrip}. Existing federated learning methods mainly discuss the scenario of clients' data heterogeneity under a balanced global class distribution, ignoring the phenomenon of an unbalanced global class distribution in actual scenarios. In these scenarios, the global data often follows a long-tailed distribution, where a large number of samples are concentrated in a few classes, while other classes are represented by only a few samples. As shown in Fig.~\ref{long-tailed and framework} (left), the classes with a large number of samples are called head class and the classes with a small number of samples are called tail class. Under the dual challenges of client data heterogeneity and the long-tailed distribution of global data, it gets harder for FL to train an effective global model, which we call federated long-tailed learning~\cite{DBLP:journals/pami/ZhangKHYF23}, ~\cite{DBLP:conf/ijcai/ShangLHW22}. For instance, multiple companies come together for joint training to develop an autonomous driving model~\cite{DBLP:conf/ivs/NguyenDTNDPTT22}. Normal driving behaviours are well represented in the dataset, while rare critical behaviours are underrepresented. This leads to local models that can handle typical situations but fail to correctly handle rare scenarios, making them perform poorly in emergencies, such as sudden obstacles, sharp turns and so on. Furthermore, insufficient recognition and prediction of different behaviours of pedestrians and non-motor vehicles will lead to the model not being able to respond correctly to sudden situations, increasing the risk of accidents. This indicates that tail class judgement is also a significant factor, and that a method to resolve the federated long-tail problem is urgently required.

To address the impact of long-tailed distribution, a straightforward solution is to apply existing solutions to data heterogeneity and long-tailed distribution to the federated long-tailed problem. However, the extensive experimental results in Table~\ref{table1} and Table~\ref{table2} show that these solutions do not lead to effective improvement. We propose a novel privacy-preserving federated learning method, FedLF, inspired by FedRS~\cite{DBLP:conf/kdd/LiZ21} and Logit adjustment~\cite{DBLP:conf/iclr/MenonJRJVK21}. It counteracts the federated long-tailed problem by adaptive logit adjustment, continuous class centred optimization, and feature decorrelation. Specifically, we make three modifications to the local training. Firstly, we adaptively adjust the logits of the local model according to the different local data distribution information of each client. The purpose is to ensure that the model does not overfit the head classes and can pay more attention to the tail classes, ensuring that the model treats each class fairly. Secondly, we maintain a set of continuously optimizable class centers locally, which greatly improves intra-class compactness and inter-class separability by adjusting the distance between features and their corresponding class centers. Finally, we continuously reduce the similarity among features by introducing the relationship matrix of features. Unlike complex algorithms that need to deal with large amounts of data or expose private data~\cite{DBLP:conf/ijcai/ShangLHW22}, ~\cite{DBLP:conf/nips/LuoCHZLF21}, our method utilizes local information to operate on local training, ensuring simplicity, efficiency, and data privacy. The contributions of this paper can be summarised as follows:

\begin{itemize}
\item We study federated learning with client-side data heterogeneity and global long-tailed distribution, where the server does not have access to data and distribution sensitive information.
\item We propose a novel privacy-preserving federated learning method FedLF to address the problem of poor tail classes performance of the model. In particular, only the local training process of the model needs to be modified to achieve satisfactory results without the risk of privacy disclosure.
\item Our method achieves superior performance through extensive comparative experiments on the benchmark datasets CIFAR-10-LT and CIFAR-100-LT with seven state-of-the-art methods.
\end{itemize}

\section{Related Work}
\subsection{Federated Learning with Data Heterogeneity}

One of the most common challenges in federated learning is data heterogeneity. During the training process of federated learning, data heterogeneity hinders the convergence speed of the model and leads to degradation of model performance. Currently, many schemes are proposed to solve the data heterogeneity problem, which are mainly classified into two strategies: client side and server side. On the client side~\cite{DBLP:conf/mlsys/LiSZSTS20}, ~\cite{DBLP:conf/icpads/JinCGL22}, strategies with local regulation limit the training process. For instance, FedProx~\cite{DBLP:conf/mlsys/LiSZSTS20} introduces regularization terms to prevent local updates from significantly deviating from the global model, thus minimizing the impact of data heterogeneity. An alternative approach is FedDyn~\cite{DBLP:conf/icpads/JinCGL22}, which incorporates server-distributed penalty terms into each client's learning objective in each round of training, thereby guiding the local model towards global optimisation. On the server side~\cite{DBLP:conf/aaai/ZhangLHC24}, global knowledge is usually used to mitigate the negative impact of data heterogeneity among clients. An excellent example is FedTGP~\cite{DBLP:conf/aaai/ZhangLHC24}, which maintains a set of trainable global prototypes on the server to help clients train prototypes with better intra-class compactness and inter-class separation. Although these methods effectively solve the challenge of data heterogeneity among clients, they generally ignore the long-tailed distribution of global data, which often leads to poor performance of the model in judging the tail classes.

\subsection{Long-tailed Learning}

Long-tailed data distributions widely exist in the real world, which put forward new requirements for the development of deep learning, have received extensive attention in research~\cite{DBLP:journals/pami/ZhangKHYF23}. The existing solutions to the long-tailed problem are mainly divided into two perspectives: data side and model side. On the data side, it mainly contains two kinds of approaches: reweighting and resampling. Reweighting methods~\cite{DBLP:conf/iccv/LinGGHD17}, ~\cite{DBLP:conf/cvpr/CuiJLSB19} modify the weights of the loss values assigned to different classes of samples. For instance, Focal Loss~\cite{DBLP:conf/iccv/LinGGHD17} assigns greater weight to challenging samples based on predicted probabilities, which enables the model to prioritize the training of difficult samples. Resampling methods~\cite{DBLP:conf/iccv/ZhangP21}, ~\cite{DBLP:conf/iccv/ZangHL21} mitigate the detrimental effect of the limited number of tail classes on the performance of the model through under-sampling the head classes or over-sampling the tail classes. On the model side, it mainly includes model decoupling and logit adjustment mechanism. Model decoupling ~\cite{DBLP:conf/iclr/KangXRYGFK20}, ~\cite{DBLP:conf/eccv/WangLKLLTHF20} focuses on reducing the model's preferences by recalibrating the classifiers, which prompts the model to look at each class more fairly. Logit adjustment~\cite{DBLP:conf/iclr/MenonJRJVK21}, ~\cite{DBLP:conf/cvpr/HongHCSKC21} is a more fine-grained solutionand aims to equalize the impact of each class, improving the model's overall accuracy. Such as LADE~\cite{DBLP:conf/cvpr/HongHCSKC21} matches the target label distribution by post-processing the model prediction trained by the cross-entropy loss and the Softmax function. However, all the above schemes assume a centralised training scenario. Faced with the challenges posed by distributed training, most of these methods are ineffective. Moreover, in the case of data heterogeneity, the complexity of training increases further, making it lose its effectiveness. Therefore, it is important to develop an approach that deals with the federated long-tailed problem.

\subsection{Federated Learning with Long-tailed Data}

To address the federated long-tailed learning problem, recent research uses mechanisms such as distillation~\cite{DBLP:conf/icmcs/ShangLCW22}, model decoupling~\cite{DBLP:conf/ijcai/ShangLHW22}, and so on. In FEDIC~\cite{DBLP:conf/icmcs/ShangLCW22}, A new distillation method with logit adjustment and a calibration gating network is proposed to alleviate the problems associated with long-tailed data. CReFF~\cite{DBLP:conf/ijcai/ShangLHW22} develops a set of constantly updated federated features to retrain classifiers on the server-side, achieving comparable performance to training models on real data. While the above methods relieve the problems posed by long tailed data to some extent, they usually require complex operations using auxiliary data or may pose significant risks to data privacy. Therefore, we propose FedLF, which requires only a modification of the local training part to obtain superior performance. In addition, it also provides maximum performance of tail classes, with details in Table~\ref{table1} and Table~\ref{table2}.

\section{Proposed Method}

Our work is inspired by FedRS and Logit adjustment. In this section, we first introduce some basic notations and then present three modifications for local training. Finally, we describe the overall optimization objective. The overall framework of FedLF is shown in Fig.~\ref{long-tailed and framework} (right).

\subsection{Preliminaries}\label{AA}

\textbf{Settings and Notations.} We consider a classical federated learning setup where $K$ clients with heterogeneous datasets $\mathcal{D}_1, \mathcal{D}_2, \mathcal{D}_3, \ldots, \mathcal{D}_K$. They collaborate on a task of $C$ classification. Our goal is to learn a global model without uploading the data to the server. In this paper, we set ${x}_{i}$ be the $i$-th input sample and  ${y}_{i}$ corresponds to the label. The global data $\mathcal{D} \triangleq \cup _k \mathcal{D}^k$ is a long-tailed distribution $X = \{({x}_{i},{y}_{i}){|}_{i=1}^{{N}_{all}}, x_i \in \mathcal D,{y}_{i}\in \{1, \ldots,C\}\}$, where ${N}_{all}$ expresses the number of all samples in $\mathcal{D}$.

We define the total number of samples for each client dataset as $N$, which is not necessarily the same for each client. $\phi $ is the distance function, $B$ is the batchsize of local training. For $i$-th sample $x_i$, we denote $h_i=F_\theta\left(x_i\right) \in \mathbb{R}^d$ as the feature vector, where $d$ is the feature dimension. In order to simplify, the bias is omitted, and the weight matrix of the last classification layer is denoted as ${W}=\left[{w}_1, {w}_2, \ldots, {w}_C\right]^{\top} \in \mathbb{R}^{C \times d}$.

\textbf{Basic Algorithm of Federated Learning.} In this paper, we use FedAvg ~\cite{DBLP:conf/aistats/McMahanMRHA17} as the foundational algorithm, upon which we propose our improvements. The typical federated learning process unfolds as follows: In round $t$, the server initially distributes the global model $\mathbf{w}^t$ to all participating clients. Each client $k$, using their unique local dataset $\mathcal{D}_k$ for $k = 1, \ldots, K$, updates their local model $\mathbf{w}_k^t$ according to the following update rule:

%%%%%% 公式应该如下写：
\begin{align}\label{client_update}
\mathbf{w}_k^{t+1} \leftarrow \mathbf{w}_k^t - \eta \nabla_{\mathbf{w}} \ell\left(\mathbf{w}^t ; \mathcal{D}_k\right),
\end{align}
 where $\eta$ denotes the learning rate, and $\ell$ denotes the loss function, typically a cross-entropy loss in classification tasks. Following the local updates, a subset of clients, denoted by ${K}^t$, is selected to upload their updated models to the server. The server then aggregates these models using a weighted averaging scheme based on the volume of data each client contributes, thereby producing a new global model for the subsequent round $t+1$:
%%%%%%% FedAvg 的服务器更新
\begin{align}\label{server_update}
\mathbf{w}^{t+1} = \sum_{k \in {K}^t} \frac{\left|\mathcal{D}_k\right|}{\sum_{k \in {K}^t}\left|\mathcal{D}_k\right|} \mathbf{w}_k^{t+1}.
\end{align}

%%%%%% 第一个组件
\subsection{Adaptive Logit Adjustment}

To address the imbalance in the distribution of classes, we locally adjust the classifier's influence weights on each class on the client side. This adjustment makes the training process more fair and improves sensitivity to all classes. Specifically, the score matrix for each class is multiplied by the adjusted local label distribution matrix. The details are as follows:

Firstly, the local label distribution is $ dist=[{n}_1, {n}_2,..., {n}_C]$, where ${n}_i$ denote the number of samples in class $i$. All elements within ${dist}$ are divided by the total number of local samples. The normalised $ndist$ is used to calculate the adjustment matrix $adist$:

\begin{align}
{adist} =\frac{{ ndist }}{\max ({ ndist })} \cdot(1.0-\alpha)+\alpha \cdot 1.
\end{align}
Here, the smoothing factor $\alpha$ is a critical hyperparameter, $1$ is the unit vector. When $\alpha$ is close to 0, the $adist$ are primarily determined by the original normalized distribution. As $\alpha$ approaches 1, the weights in $adist$ for each class become nearly equal. This smoothing step introduces a certain degree of uniformity while preserving the original distributional characteristics of the data. It ensures that the model does not become overly sensitive to classes with extreme distribution during training. By retaining the essential information of each class and reducing the influence of outliers, this approach enhances the model's ability to learn from less frequent classes. Ultimately, this helps improve the model's generalization ability and fairness across uneven datasets.

Secondly, the calculated $adist$ is element-wise multiplied with the original score matrix ${h}_i \cdot{W}^{\top}$to obtain the adjusted logits ${z}_i$:

\begin{align}
{z}_i = {adist} \odot {h}_i \cdot {W}^{\top}.
\end{align}

Finally, the loss function ${L_A}$ is calculated from the modified logits $z_{i, j}$:

\begin{align}
L_A=-\frac{1}{N} \sum_{i=1}^N y_i\log \left(\frac{\exp \left(z_{i, y_i}\right)}{\sum_{j=1}^C \exp \left(z_{i, j}\right)}\right),
\label{L_A}
\end{align}
where $z_{i, j}$ denotes the adjusted logit of sample $x_i$ on class $j$, $N$ denoted total number of sample.

\subsection{Class Center Optimization}

Inspired by contrast loss, we adjusting the Euclidean distance of features and the corresponding class centers, which improve  the model’s ability to discriminate among samples of different classes. Particularly, we maintain constantly updated set of class centers locally $\hat{P} = [\hat{{p}_{c}}\mid c=1,2,...,C]$. We assume that ${h}_{i}^{c}$ represents the features whose sample $x_i$ is class $c$, ${N}_{c}$ is the total number of local samples belonging to class $c$, and $\hat{{p}_{c}}$ is computed as follows:

\begin{align}
\hat{{p}_{c}}=\frac{1}{N_c}\displaystyle\sum_{i=1}^{N_c} {h}_{i}^{c},
\end{align}
and then, we keep optimizing the class centers $\hat{P}$ during the training process by:

\begin{align}
L_C= \displaystyle  \displaystyle\sum_{c=1}^{C}\sum_{i=1}^{N_c}-\log(\frac{{e}^{-\phi ({h}_{i}^{c},\hat{{p}_{c}})}}{{e}^{-\phi ({h}_{i}^{c},\hat{{p}_{c}})}+{\sum}_{{c}^{'},{c}^{'}\neq c}{e}^{-\phi({h}_{i}^{c},\hat{{p}_{{c}^{'}}})}}).
\label{L_C}
\end{align}
Here, ${c}^{'}$ represents all classes not equal to class $c$. Although the above formula is the standard contrast loss, it does not significantly reduce the intra-class distance, and the learned inter-class boundaries are not clear enough. 

To further clarify the boundaries, we force the learning of inter-class margins $Q$ during training by introducing a class spacing threshold $\tau$, which is calculated as follows:

\begin{align}
Q=\min(({\max_{c\subseteq [C], {c}^{'}\subseteq [C], c\ne c^{'}}{\phi (\hat{p}_{c},\hat{p}_{{c}^{'}})}}),\tau),
\end{align}
where $\tau$ is a hyperparameter. The loss function is rewritten as follows:

\begin{align}
L_C= \displaystyle\sum_{c=1}^{C} \sum_{i=1}^{N_C}-(\log(\frac{{e}^{-\phi ({h}_{i}^{c},\hat{{p}_{c}})+Q)}}{{e}^{-(\phi ({h}_{i}^{c},\hat{{p}_{c}})+Q)}+{\sum}_{{c}^{'},{c}^{'}\neq c}{e}^{-\phi({h}_{i}^{c},\hat{{p}_{{c}^{'}}})}}).
\end{align}

The $L_C$ partially takes into account the distribution of sample features, which helps the model to better distinguish among samples of different classes and improves the classification performance of the model. The introduction of the class spacing threshold controls the spacing among classes to avoid the influence of abnormal samples and enhances the generalization ability of the model.

%%%%%%%%%%%第三个组件
\subsection{Feature Decorrelation}

To further enhance the robustness of the model against correlated feature distributions, we introduce the loss function for decorrelation. Through evaluating the covariance matrix of the features and penalising non-diagonal elements , this loss function aims to minimise the correlation among features. The method is as follows: 

The feature matrix $\mathbf{X}$ is first normalised:

\begin{align}
\mathbf{X}_{i j}^{\text {norm }}=\frac{\mathbf{X}_{i j}-\mu_j}{\sigma_j},
\end{align}
Here $\mathbf{X}_{i j}^{\text {norm }}$ is the element in row $i$ and column $j$. $\mu_j$ and $\sigma_j$ represent the mean and standard deviation of column $j$. Then estimate the correlation matrix $\mathbf{Cor}$:

\begin{align}
\mathbf{Cor}=\frac{1}{B} \mathbf{X^{norm}}^T \mathbf{X^{norm}},
\end{align}
where $X^{norm}$ is the normalised representation matrix, $B$ is the batch size. The loss function $L_D$ is computed as follows:
%%%%  第三个组件的公式

\begin{align}
L_D=\sum_{i=1}^N \sum_{j=1}^N\left(\mathbf{Cor}_{i j}\right)^2.
\label{L_D}
\end{align}
Here, ${Cor}_{i j}$ is the element of row $i$ and column $j$ of the correlation matrix $Cor$. The non-diagonal (${Cor}_{i j},i \ne j$) elements within the matrix are the values corresponding to the correlation among the features. We effectively reduce the correlation among the features by continuously optimizing the loss function, enhance the robustness and generalization of the model.

\subsection{Overall Optimization Objective}

We combine the above three losses, and this combined loss function optimizes intra-class compactness and inter-class separation while adapting to logits adjustment, and improves feature independence. It significantly improves the generalization ability and effectiveness of the overall model. The combine loss are as follows:

\begin{align}
 L = L_A + \lambda  L_C + \gamma  L_D,
\end{align}
Here, $\lambda$ and $\gamma$  are hyperparameters, and the clients trains using $L$. Algorithm~\ref{fedad} summarizes the recommendation framework FedLF.

%  整体的框架图

\begin{algorithm2e}
\footnotesize % 或者使用 \footnotesize
\caption{FedLF}
\label{fedad}
\DontPrintSemicolon
\LinesNumbered
\KwIn{Initialized global model $\bw^0$, smoothing factor $\alpha$, class spacing threshold $\tau$, weights of the two loss functions $\lambda$ and $\gamma$.}
\BlankLine
\KwOut{global model $\bw^{t+1}$.}
\BlankLine
\For{$t=1$ \KwTo $T$}{
    %\tcp{Server executes:}
		Randomly select a set of active clients ${K}^t$.\\
		\tcp{Clients execute:}
		\For{$k\in {K}^t$}{	
                %%\ref{client_update 用于引用文中的标签
                Computer $L_A$ by Equation~\ref{L_A};\\
                Computer $L_C$ by Equation~\ref{L_C};\\
                Computer $L_D$ by Equation~\ref{L_D};\\
                
                Update local model $\bw_k^{t+1}$ by Equation~\ref{client_update};\\
			Send $\bw_k^{t+1}$ to the server.\\
		}
  
		\tcp{Server executes:}
		Aggregate local models to $\bw^{t+1}$ by Equation~\ref{server_update}.\\
		%Send $\bw^{t+1}=\{\bu^{t+1}, \bv^{t+1}\}$ and $\widehat{\bw}^{t+1}=\{\bu^{t+1}, \widehat{\bv}^{t+1}\}$ to clients;\\
}
\end{algorithm2e}

\section{Experimental Results}

In this section, we compare FedLF with seven state-of-the-art methods to demonstrate that FedLF is effective in relieving the federated long-tailed problem. In order to evaluate the effectiveness of all methods in more detail, we conduct extensive experiments on two benchmark datasets and evaluated four accuracies, namely head classes accuracy, middle classes accuracy, tail classes accuracy and overall accuracy.

\subsection{Experimental Setup}

The basic experimental setup is as follows: the total number of clients $K$ is set to 20, the number of local iterations $E$ for each client is 5, and the local batchsize $B$ is 32.

% 迪利克雷系数：Dirichlet concentration degree
\textbf{Implementation details.} All experiments use ResNet-8~\cite{DBLP:conf/cvpr/HeZRS16} as the base model and run under PyTorch with an Nvidia GeForce RTX 3060 Laptop GPU. We conduct experiments on the CIFAR-10/100-LT~\cite{Krizhevsky2009LearningML} dataset and set IF~\cite{DBLP:conf/nips/CaoWGAM19} to 100, 50,and 10. IF is long-tailed factor to describe the degree of imbalance in the long-tailed case. The heterogeneous dataset is partitioned into each client using Dirichlet coefficients based on previous research~\cite{DBLP:journals/corr/abs-2401-08977}, where Dirichlet coefficients are set to 0.5 and 1.0. The learning rate is mildly set to 0.1. We set the online rate of clients to $40\%$, and the clients online each time are randomly selected. 

In addition, we have four different metrics when evaluating model performance: header classes accuracy, middle classes accuracy, tail classes accuracy, and all classes accuracy. To define the head and tail classes, we introduce thresholds. We set the classes with more samples above the thresholds as head classes, and the classes with fewer samples than the thresholds as tail classes. We have different thresholds for different degrees of long-tailed. Specifically, for CIFAR-10-LT, the thresholds are set to (1500, 200) when the long-tail factor IF is 100 and 50, and (1500, 600) when the IF is 10. For CIFAR-100-LT, the threshold is set to (200,20) when the long-tail factor IF is 100 and 50, and (300, 60) when the IF is 10.

\textbf{Baselines.} We compare with the following baseline: FedAvg~\cite{DBLP:conf/aistats/McMahanMRHA17}, FedBN~\cite{DBLP:conf/iclr/LiJZKD21}, FedRS~\cite{DBLP:conf/kdd/LiZ21}, FEDIC~\cite{DBLP:conf/icmcs/ShangLCW22}, Focal Loss~\cite{DBLP:conf/iccv/LinGGHD17}, FedProx~\cite{DBLP:conf/mlsys/LiSZSTS20}, CReFF~\cite{DBLP:conf/ijcai/ShangLHW22}.

\textbf{Hyperparameters.} Unless stated otherwise, most hyperparameters of these baseline are configured according to the original literature. We utilize the official open-source codes of these methods. There are four hyperparameters in FedLF, namely smoothing coefficient $\alpha$, class spacing threshold $\tau$, and the weights of the loss function $\lambda$, $\gamma$. The smoothing coefficient $\alpha$ is set to 0.25 with reference to~\cite{DBLP:conf/kdd/LiZ21}. It enhances the generalization ability of the model, reduces the sensitivity to outliers, and effectively reduces the interference of extreme sample distributions on model training. We set the threshold $\tau$ to 100 with reference to~\cite{DBLP:conf/aaai/ZhangLHC24}. In our loss function formulation: $L = L_A + \lambda L_C + \gamma L_D$, the hyperparameters $\lambda$ and $\gamma$ are crucial for balancing the contributions of the losses $L_C$ and $L_D$ to the aggregate loss $L$. We set both $\lambda$ and $\gamma$ to 0.01, a decision driven by the need to ensure that the additional terms enhance the model's performance without overwhelming the primary loss, $L_A$. The values of $\lambda$ and $\gamma$ are determined through a systematic exploration of various settings, where 0.01 emerged as the optimal value that subtly integrates the corrective effects of $L_C$ and $L_D$, improving overall model robustness and accuracy. For details, please see Fig.~\ref{lambda and gamma curve}.

\subsection{Results and Analysis}

The results are shown in Table~\ref{table1} ($\alpha = 0.5$) and Table~\ref{table2} ($\alpha = 1$). In the table, results in bold and underlined represent the best and second best results for that column. Our method achieves superior results in all experiments. Compared to the baseline method FedAvg, our method achieves the highest performance improvement of 13.27\% when IF=100, 
$\alpha=0.5$ at CIFAR-10-LT. This demonstrates that our method has good generalization ability in the case of severe long-tailed global class distributions.

FedProx, FedBN and FedRS mainly address data heterogeneity. The results of FedProx and FedBN perform similarly to FedAvg because they primarily focus on addressing data heterogeneity without considering long tailed distribution. In addition, FedRS, although not specifically designed for long-tailed distribution, still performs well in long-tailed cases, suggesting that logit adjustment effectively addresses the long-tailed problem.

CReFF and FEDIC consider long-tailed distribution and show good results in solving this problem. In few cases, the CReFF method slightly outperforms our method, but when identifying tail classes, our method significantly outperforms CReFF.

\begin{figure}[htp]
\begin{center}
\includegraphics[width=0.82\textwidth]{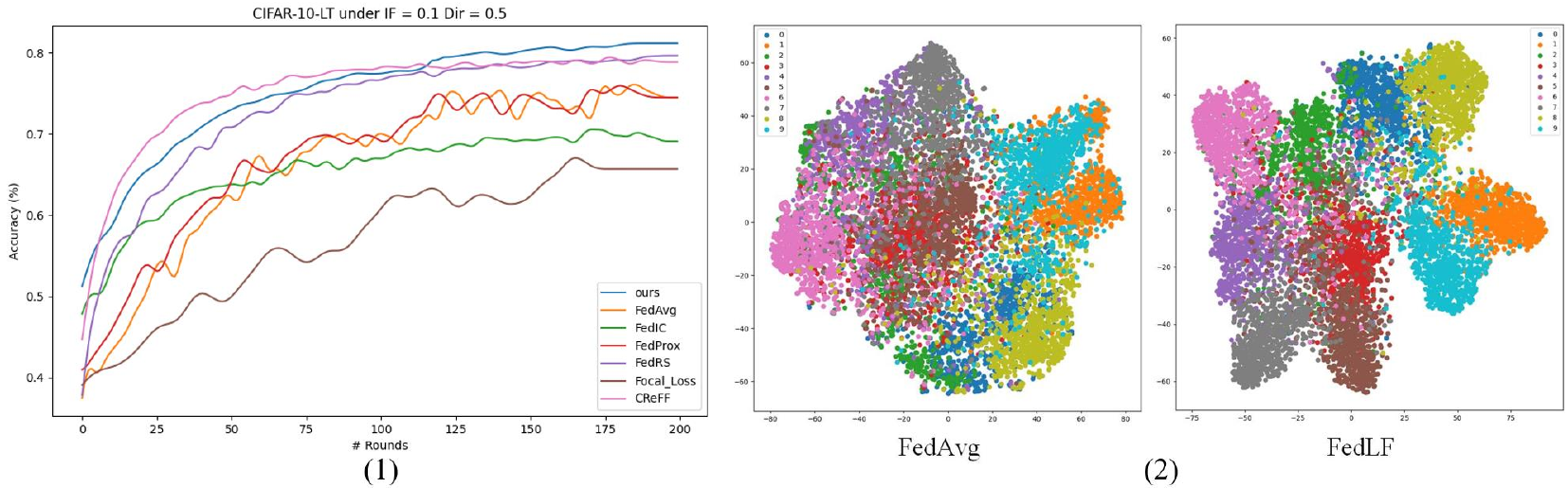}
\caption{(1) denotes the convergence curves on CIFAR-10-LT with IF=10 and $\alpha=0.5$, (2) denotes feature TSNE plot after 200 rounds of training.}
\label{convergence_curve and tsne}
\end{center}
\end{figure}

The performance gap between the above methods and ours is due to the fact that FedProx, FedBN and FedRS mainly solve the problem of data heterogeneity among clients and do not take into account the long-tailed factor, which is the most common factor in real life. CReFF and FEDIC is designed for the long-tailed problem have good overall performance but do not effectively improve the accuracy of determining the tail classes. In contrast, our approach takes into account the long-tailed problem. It obtains good overall performance without losing the ability to judge tail classes. Fig.~\ref{convergence_curve and tsne} represents the convergence curves of algorithms for the IF = 10 and $\alpha =  0.5$ and feature TSNE plot after 200 rounds of training.

%%%%% 收敛曲线图

%%%   table不带星号指的是单列
%%%%%   textwidth  才是缩放大小适应两列  原来那个是单列
\begin{table*}[!t]
\tiny
\caption{test accuracy ($\%$) by compared FL methods on CIFAR-10/100-LT at $\alpha$=0.5.}
\centering
\label{table1}
\resizebox{\textwidth}{!}{%
\begin{tabular}{llllllllllllllllllllllllll}
\hline
\multirow{3}{*}{Datsset} & Non-IID & \multicolumn{12}{c}{$\alpha=0.5$} \\ \cline{3-14} 
 & Imbalance Factor & \multicolumn{4}{c}{IF=100} & \multicolumn{4}{c}{IF =50} & \multicolumn{4}{c}{IF=10} \\ \cline{3-14} 
 & Method/Model & Head & Middle  & Tail  & All & Head & Middle & Tail & All & Head & Middle & Tail & All \\ \hline
\multirow{8}{*}{CIFAR-10-LT} & \multicolumn{1}{l|}{FedAvg} & \underline{86.17} & 53.67 & 28.07 & \multicolumn{1}{l|}{55.74} & 90.87 & 55.84 & 31.70 & \multicolumn{1}{l|}{61.52} & \textbf{85.34} & 63.30 & 64.90 & 74.48 \\
 &\multicolumn{1}{l|}{FedProx} & 83.33 & 59.60 & 28.27 & \multicolumn{1}{l|}{57.32} & \textbf{91.53} & 58.52 & 31.80 & \multicolumn{1}{l|}{63.08} & \underline{84.86} & 70.70 & 53.60 & 76.07 \\
 & \multicolumn{1}{l|}{CReFF} & 82.30 & 50.48 & 18.70 & \multicolumn{1}{l|}{\textbf{69.21}} & 89.13 & 59.30 & 31.50 & \multicolumn{1}{l|}{71.90} & 84.38 & 73.22 & 57.20 & \underline{78.51} \\
 & \multicolumn{1}{l|}{FedBN} & \textbf{86.47} & 50.35 & 29.20 & \multicolumn{1}{l|}{54.84} & \underline{91.50} & 54.30 & 36.30 & \multicolumn{1}{l|}{61.86} & 84.44 & 62.48 & 65.60 & 73.77 \\
 & \multicolumn{1}{l|}{FEDIC} & 62.40 & 60.90 & \textbf{78.03} & \multicolumn{1}{l|}{66.49} & 68.97 & 63.88 & 71.03 & \multicolumn{1}{l|}{67.55} & 73.50 & 61.40 & 73.76 & 71.21 \\
 & \multicolumn{1}{l|}{FedRS} & 65.17 & \textbf{69.05} & 67.87 & \multicolumn{1}{l|}{67.53} & 70.73 & \textbf{70.30} & \underline{79.85} & \multicolumn{1}{l|}{\underline{72.34}} & 72.54 & \textbf{83.40} & \textbf{86.00} & 78.51 \\
 & \multicolumn{1}{l|}{Focal Loss} & 80.97 & 50.95 & 13.23 & \multicolumn{1}{l|}{48.64} & 88.73 & 47.36 & 17.60 & \multicolumn{1}{l|}{53.82} & 76.76 & 65.42 & 64.50 & 71.00 \\
 & \multicolumn{1}{l|} {\cellcolor[HTML]{EFEFEF}FedLF} & \cellcolor[HTML]{EFEFEF}72.73 & \cellcolor[HTML]{EFEFEF}\underline{65.67} &\cellcolor[HTML]{EFEFEF} \underline{69.67} &\multicolumn{1}{l|}{\cellcolor[HTML]{EFEFEF} \underline{69.01}} & \cellcolor[HTML]{EFEFEF}74.43 & \cellcolor[HTML]{EFEFEF}\underline{68.86} & \cellcolor[HTML]{EFEFEF}\textbf{82.15} & \multicolumn{1}{l|}{\cellcolor[HTML]{EFEFEF}\textbf{73.19}} &\cellcolor[HTML]{EFEFEF} 80.38 &\cellcolor[HTML]{EFEFEF} \underline{78.45} &\cellcolor[HTML]{EFEFEF} \underline{85.90} &\cellcolor[HTML]{EFEFEF} \textbf{80.16} \\ \hline
\multirow{8}{*}{CIFAR-100-LT} 
 & \multicolumn{1}{l|} {FedAvg} & \textbf{67.15} & 32.10 &3.67 &\multicolumn{1}{l|}{30.58} & \underline{65.25} & 31.53 &7.50 &\multicolumn{1}{l|}{35.30} & \textbf{66.77} & 41.51 & 12.25 & 44.73 \\
 &\multicolumn{1}{l|}{FedProx} & \underline{66.90} & 33.24 & 6.10 & \multicolumn{1}{l|}{31.83} & 64.88 & 31.84 & 9.11 & \multicolumn{1}{l|}{35.68} & \underline{65.09} & 42.96 & 17.13 & 45.46 \\
 & \multicolumn{1}{l|}{CReFF} & 63.12 & 47.83 &9.00 & \multicolumn{1}{l|}{\textbf{34.60}} & \textbf{65.34} & \textbf{48.06} & 10.32 &\multicolumn{1}{l|}{ \underline{37.64}} & 60.35 & \textbf{53.74} & 15.00 &\underline{47.08} \\
 & \multicolumn{1}{l|}{FedBN} & 62.65 &30.06 & 5.03 & \multicolumn{1}{l|}{29.07} &60.79 & 28.91 & 4.78 & \multicolumn{1}{l|}{32.22} & 55.68 & 42.04 & 13.00 & 42.72 \\
 & \multicolumn{1}{l|} {FEDIC} &50.30 & 41.26 & \underline{10.49} &\multicolumn{1}{l|}{33.67} & 48.13 & 39.08 & 13.00 & \multicolumn{1}{l|}{36.74} & 57.84 & 40.16 & \underline{17.65} & 41.93 \\
 & \multicolumn{1}{l|}{FedRS} & 50.45 &\underline{41.80} & 8.53 &\multicolumn{1}{l|}{33.23} & 48.79 & 40.45 & \underline{15.44} & \multicolumn{1}{l|}{37.12} & 51.27 & 48.69 & 15.88 & 46.70 \\
 & \multicolumn{1}{l|}{Focal Loss} & 58.70 & 28.86 & 2.23 & \multicolumn{1}{l|}{26.84} &59.17 &27.10 & 5.44 & \multicolumn{1}{l|}{30.94} & 60.14 & 41.39 & 12.63 & 43.21 \\
 & \multicolumn{1}{l|}{\cellcolor[HTML]{EFEFEF}FedLF} &\cellcolor[HTML]{EFEFEF} 52.10 &\cellcolor[HTML]{EFEFEF} \textbf{41.86} &\cellcolor[HTML]{EFEFEF} \textbf{11.33} &\multicolumn{1}{l|}{\cellcolor[HTML]{EFEFEF} \underline{34.48}} &\cellcolor[HTML]{EFEFEF} 49.67 &\cellcolor[HTML]{EFEFEF} \underline{42.41} & \cellcolor[HTML]{EFEFEF}\textbf{16.06} & \multicolumn{1}{l|}{\cellcolor[HTML]{EFEFEF}\textbf{39.52}} &\cellcolor[HTML]{EFEFEF} 53.18 & \cellcolor[HTML]{EFEFEF}\underline{50.63} &\cellcolor[HTML]{EFEFEF} \textbf{19.38} & \cellcolor[HTML]{EFEFEF}\textbf{48.69} \\ \hline
\end{tabular}%
}
\end{table*}

%%% alpha = 1
\begin{table*}[!t]
\tiny
\caption{Test accuracy ($\%$) by compared FL methods on CIFAR-10/100-LT at $\alpha$=1.0.}
\centering
\label{table2}
\resizebox{\textwidth}{!}{%
\begin{tabular}{llllllllllllllllllllllllll}
\hline
& Non-IID & \multicolumn{12}{c}{$\alpha=1$} \\ \cline{3-14} 
 & Imbalance Factor & \multicolumn{4}{c}{IF=100} & \multicolumn{4}{c}{IF =50} & \multicolumn{4}{c}{IF=10} \\ \cline{3-14} 
\multirow{-3}{*}{Datsset} & Method/Model & Head & Middle  & Tail  & All & Head & Middle & Tail & All & Head & Middle & Tail & All \\ \hline
 & \multicolumn{1}{l|}{FedAvg} & \underline{83.93} & 61.28 & 35.17 & \multicolumn{1}{l|}{60.24} & \underline{91.93} & 60.84 &22.60 &\multicolumn{1}{l|}{62.52} & \textbf{87.76} & 69.07 &47.20 & 76.23 \\
 & \multicolumn{1}{l|}{FedProx} & 83.03 & 65.87 & 35.40 & \multicolumn{1}{l|}{61.88} & 90.13 & 61.78 & 28.45 & \multicolumn{1}{l|}{63.62} & 85.02 & 73.02 & 65.50 & 78.27 \\
 &\multicolumn{1}{l|}{CReFF} & \textbf{89.17} & 52.08 & 23.50 & \multicolumn{1}{l|}{\textbf{69.94}} & \textbf{93.60} & 55.34 & 25.65 & \multicolumn{1}{l|}{\underline{72.68}} & \underline{86.46} &71.95 & 60.50 & 80.56 \\
 & \multicolumn{1}{l|}{FedBN} & 80.00 &60.65 & 36.70 & \multicolumn{1}{l|}{59.27} &79.90 & 58.60 & 35.65 & \multicolumn{1}{l|}{61.68} & 73.48 & \underline{82.94} & 62.20 & 76.14 \\
 & \multicolumn{1}{l|}{FEDIC} & 73.37 & 59.35 & 68.50 & \multicolumn{1}{l|}{66.30} &71.40  & 59.15 &71.60  &\multicolumn{1}{l|}{69.05} & 73.43 & 68.00 & 75.94 & 73.16 \\
 & \multicolumn{1}{l|}{FedRS} &62.27 & \underline{69.45} & \underline{69.50} & \multicolumn{1}{l|}{67.31} &71.43 & \underline{70.80} & \underline{77.30} & \multicolumn{1}{l|}{72.29} & 79.08 & 80.95 & \textbf{90.90} & \underline{80.93} \\
 & \multicolumn{1}{l|}{Focal Loss} & 74.53 &56.63 &21.10 & \multicolumn{1}{l|}{51.34} &85.40 & 51.64 & 9.30 & \multicolumn{1}{l|}{53.30} & 82.54 & 66.90 & 66.60 &73.69 \\
\multirow{-8}{*}{CIFAR-10-LT} & \multicolumn{1}{l|}{\cellcolor[HTML]{EFEFEF}FedLF} & \cellcolor[HTML]{EFEFEF}65.27 & \cellcolor[HTML]{EFEFEF}\textbf{69.87} & \cellcolor[HTML]{EFEFEF}\textbf{70.77} & \multicolumn{1}{l|}{\cellcolor[HTML]{EFEFEF}\underline{68.76}} & \cellcolor[HTML]{EFEFEF}74.17 & \cellcolor[HTML]{EFEFEF}\textbf{71.92} & \cellcolor[HTML]{EFEFEF}\textbf{80.25} & \multicolumn{1}{l|}{\cellcolor[HTML]{EFEFEF}\textbf{74.05}} & \cellcolor[HTML]{EFEFEF}81.60 & \cellcolor[HTML]{EFEFEF}\textbf{82.95} & \cellcolor[HTML]{EFEFEF}\underline{86.60} & \cellcolor[HTML]{EFEFEF}\textbf{82.64} \\ \hline
 & \multicolumn{1}{l|}{FedAvg} & \underline{66.95} &33.88 &3.87 &\multicolumn{1}{l|}{31.18} & \textbf{67.33} & 32.03 &6.78 & \multicolumn{1}{l|}{35.96} & \underline{66.86} & 45.33 & 14.25 &46.33 \\
 & \multicolumn{1}{l|}{FedProx} & 66.85 &33.64 &3.83 & \multicolumn{1}{l|}{31.35} & \underline{66.92} &31.88 & 9.83 & \multicolumn{1}{l|}{36.32} & 66.50 & 46.93 &\textbf{16.00} &47.15 \\
 & \multicolumn{1}{l|}{CReFF} & \textbf{67.23} & \textbf{55.87} & 3.20 & \multicolumn{1}{l|}{\textbf{35.23}} & 65.89 & 53.62 & 14.32 &\multicolumn{1}{l|}{\underline{38.65}}  & \textbf{70.83} & \textbf{58.01} & 15.06 & \underline{48.03} \\
 & \multicolumn{1}{l|}{FedBN} & 57.10 & 32.94 & 3.10 & \multicolumn{1}{l|}{29.28} & 60.83 & 28.81 & 9.72 & \multicolumn{1}{l|}{33.06} & 60.18 & 41.40 & 10.13 & 43.03 \\
 & \multicolumn{1}{l|}{FEDIC} & 49.85 & 41.36 &\underline{8.60} & \multicolumn{1}{l|}{33.98} & 50.63 & 40.19 &\underline{11.68} & \multicolumn{1}{l|}{37.26} & 65.39 & 42.60 & 13.03 & 45.36 \\
 & \multicolumn{1}{l|}{FedRS} & 51.15 &43.10 & 8.37 & \multicolumn{1}{l|}{34.29} &46.54 & \underline{41.48} & 17.00 & \multicolumn{1}{l|}{37.90} &48.73 & 50.89 & 12.50 & 47.34 \\
 & \multicolumn{1}{l|}{Focal Loss} & 60.55 & 28.78 &3.30 & \multicolumn{1}{l|}{27.49} & 60.21 & 29.83 &2.33 & \multicolumn{1}{l|}{32.17} & 64.55 & 41.64 &12.12 & 44.32 \\
\multirow{-8}{*}{CIFAR-100-LT} & \multicolumn{1}{l|}{\cellcolor[HTML]{EFEFEF}FedLF} & \cellcolor[HTML]{EFEFEF}49.55 & \cellcolor[HTML]{EFEFEF}\underline{44.54} & \cellcolor[HTML]{EFEFEF}\textbf{9.93} & \multicolumn{1}{l|}{\cellcolor[HTML]{EFEFEF}\underline{35.16}} & \cellcolor[HTML]{EFEFEF}52.17 & \cellcolor[HTML]{EFEFEF}\textbf{42.19} & \cellcolor[HTML]{EFEFEF}\textbf{13.83} & \multicolumn{1}{l|}{\cellcolor[HTML]{EFEFEF}\textbf{40.12}} & \cellcolor[HTML]{EFEFEF}54.27 & \cellcolor[HTML]{EFEFEF}\underline{52.31} & \cellcolor[HTML]{EFEFEF}\underline{15.12} & \cellcolor[HTML]{EFEFEF}\textbf{49.77 }\\ \hline
\end{tabular}%
}

\end{table*}

\subsection{Ablation Study}

\begin{table*}[!t]
\tiny  % 或者使用 
\caption{ Ablation Experiments result($\%$) for IF=50 and IF=10 at $\alpha=0.5$}
\centering
\label{table3}
% \resizebox{\columnwidth}{!}{%
\begin{tabular}{c|cc|cccc}
\hline
\multirow{2}{*}{Long-tailed factor} & \multicolumn{2}{c|}{Components} & \multicolumn{4}{c}{Multi-precision} \\ \cline{2-7} 
 & $L_C$ & $L_D$ & Head & Middle & Tail & All \\ \hline
\multirow{4}{*}{IF = 50} &\ding{55}  &\ding{55}  & 72.70 & 68.34 & 72.10 & 70.40 \\
 & \ding{51} &\ding{55}  & 73.63 & 68.34 & 79.20 & 72.10 \\
 & \ding{55} & \ding{51} & 73.13 & 69.70 & 77.80 & 72.35 \\
 & \ding{51} & \ding{51} & 74.43 & 68.86 & 82.15 & 73.19 \\ \hline
\multirow{4}{*}{IF = 10} &\ding{55}  &\ding{55}  & 71.64 & 83.53 & 89.80 & 78.21 \\
 & \ding{51} & \ding{55} & 73.96 & 82.75 & 89.70 & 79.05 \\
 & \ding{55} & \ding{51} & 74.92 & 81.13 & 91.10 & 79.02 \\
 & \ding{51} & \ding{51} & 77.48 & 81.80 & 88.20 & 80.28 \\ \hline
\end{tabular}%
%}
\end{table*}

%% 下面是rebuttal阶段补充的消融实验

% Please add the following required packages to your document preamble:
% \usepackage{graphicx}
\begin{table*}[!t]
\tiny  % 或者使用 \footnotesize
\caption{ Ablation Experiments result($\%$) on CIFAR-10-LT at $\alpha$=0.1 and IF=0.1}
\centering
\label{table4}
%\small
\begin{tabular}{c|cccccc}
\hline
Non-IID and IF factor& \multicolumn{4}{c}{$\alpha$ = 0.1 and IF = 0.1} \\ \hline
Method/Model & Many & Middle & Tail & All \\
FedAvg & 75.80 & 45.30 & 45.00 & 60.52 \\
FedProx & \textbf{79.32} & 62.52 & 43.60 & 69.03 \\
CReFF & \underline{75.98} & 44.55 & 42.30 & \underline{73.56} \\
FedBN & 57.80 & 50.08 & \textbf{88.90} & 57.82 \\
FEDIC & 74.60 & 54.23 & 46.90 & 72.16 \\
FedRS & 72.12 & \underline{63.30} & 65.50 & 67.93 \\
Focal Loss & 65.36 & 61.180 & 43.60 & 61.51 \\
\cellcolor[HTML]{EFEFEF} FedLF & \cellcolor[HTML]{EFEFEF}75.70 & \cellcolor[HTML]{EFEFEF}\textbf{72.20} & \cellcolor[HTML]{EFEFEF}\underline{69.90} & \cellcolor[HTML]{EFEFEF}\textbf{73.72} \\ \hline
\end{tabular}%
%}
\end{table*}

% 每个参数权重是否敏感
Our loss function consists of three components, in order to assess the impact of each component of FedLF on model performance in the same long-tailed environment, we design a series of ablation experiments. By progressively removing or modifying weights of components, we aim to clarify the contribution of each component to the overall performance. In particular, we hope to reveal the role of each component in improving model accuracy under data distributions characterized by varying degrees of long-tailed.

\textbf{Comparisons under the same long-tailed.} We evaluate the impact of each component in the same long-tailed environment to demonstrate its effectiveness.

\begin{itemize}
\item Effectiveness of $L_C$: The effect of $L_C$ is evident in Table~\ref{table3}. The experimental results are significantly improved with the addition of this component, which increases the ability of intra-class compactness and inter-class separability of features.

\item Effectiveness of $L_D$: As can be seen from Table~\ref{table3}, $L_D$ plays an vital role. After adding this component, the experimental results are significantly improved, indicating that this loss function effectively reduces the correlation of features.

\end{itemize}

\textbf{Comparisons under the different long-tailed and heterogeneous.} 
As shown in Table~\ref{table1} and Table~\ref{table2}, varying degrees of long-tailed distribution significantly impacts the algorithm's accuracy. As the extent of the long-tailed distribution increases, the accuracy of the algorithm correspondingly decreases. To ensure the rigor of our conclusions, we conduct ablation experiments under different levels of long-tail distribution to verify the effectiveness of each component. The experimental results are shown in Table~\ref{table3}, indicating that each component is effective under different degrees of long-tailed environmental conditions. In addition, we conduct an experiment in a serious heterogeneous scenario, such as $\alpha$=0.1 in Table~\ref{table4}. The results show that FedLF demonstrates effectiveness in such severe heterogeneous scenarios.

\begin{figure}[htp]
\begin{center}
\scalebox{0.35}{  % 将图片缩小到原始大小的70%
    \includegraphics{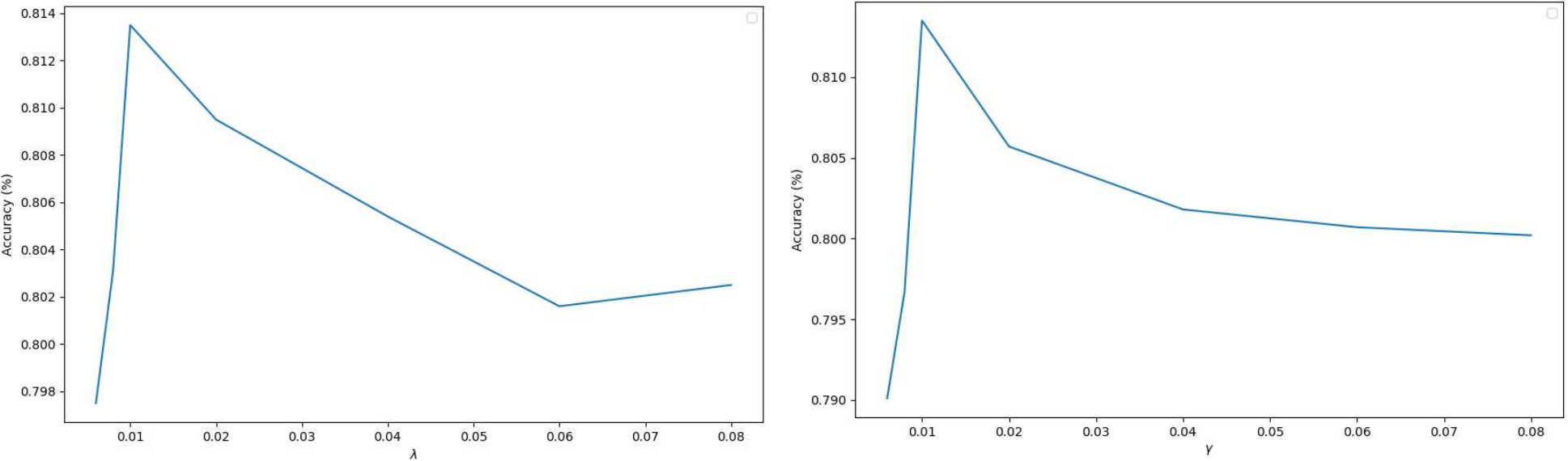}
}
\caption{Impact of $\lambda$ and $\gamma$ weights.}
\label{lambda and gamma curve}
\end{center}
\end{figure}

\textbf{Impact of different weights.}
We investigate the effect of the weights $\lambda$ of $L_C$ and $\gamma$ of $L_D$ on model performance. The 3 show that different parts of the loss function play different roles in training and optimization. Therefore, to better exploit the contribution of each loss term, we adjust the weights of $L_C$ and $L_D$ separately and observe their effects on model performance. The results are shown in Fig.~\ref{lambda and gamma curve}. The model performance is optimal when $\lambda$ and $\gamma$ are set to 0.01.

\section{Conclusion}

In this paper, we present FedLF to enhance federated learning under data heterogeneity and global long-tailed data. FedLF is a client-side approach, which introduces three modifications in the local training phase: adaptive logits adjustment, continuous class centred optimization, and feature decorrelation. Furthermore, the effectiveness of each component of FedLF is verified. Experiments show that FedLF achieves superior results on datasets with heterogeneous and long-tailed settings compares to seven other state-of-the-art methods. For future works, we aim to study federated learning robustness against noisy labels \cite{fednoro,fedlsr,fnbench,fedelc}, especially in long-tailed data environments.

%\acks{Acknowledgements should go at the end, before appendices and references. You can uncomment this for the camera-ready version on paper acceptance.}

%\bibliographystyle{plain}
\bibliography{acml24}

\begin{thebibliography}{35}
\providecommand{\natexlab}[1]{#1}
\providecommand{\url}[1]{\texttt{#1}}
\expandafter\ifx\csname urlstyle\endcsname\relax
  \providecommand{\doi}[1]{doi: #1}\else
  \providecommand{\doi}{doi: \begingroup \urlstyle{rm}\Url}\fi

\bibitem[Cao et~al.(2019)Cao, Wei, Gaidon, Ar{\'{e}}chiga, and Ma]{DBLP:conf/nips/CaoWGAM19}
Kaidi Cao, Colin Wei, Adrien Gaidon, Nikos Ar{\'{e}}chiga, and Tengyu Ma.
\newblock Learning imbalanced datasets with label-distribution-aware margin loss.
\newblock In Hanna~M. Wallach, Hugo Larochelle, Alina Beygelzimer, Florence d'Alch{\'{e}}{-}Buc, Emily~B. Fox, and Roman Garnett, editors, \emph{Advances in Neural Information Processing Systems 32: Annual Conference on Neural Information Processing Systems 2019, NeurIPS 2019, December 8-14, 2019, Vancouver, BC, Canada}, pages 1565--1576, 2019.
\newblock URL \url{https://proceedings.neurips.cc/paper/2019/hash/621461af90cadfdaf0e8d4cc25129f91-Abstract.html}.

\bibitem[Cui et~al.(2019)Cui, Jia, Lin, Song, and Belongie]{DBLP:conf/cvpr/CuiJLSB19}
Yin Cui, Menglin Jia, Tsung{-}Yi Lin, Yang Song, and Serge~J. Belongie.
\newblock Class-balanced loss based on effective number of samples.
\newblock In \emph{{IEEE} Conference on Computer Vision and Pattern Recognition, {CVPR} 2019, Long Beach, CA, USA, June 16-20, 2019}, pages 9268--9277. Computer Vision Foundation / {IEEE}, 2019.
\newblock \doi{10.1109/CVPR.2019.00949}.
\newblock URL \url{http://openaccess.thecvf.com/content\_CVPR\_2019/html/Cui\_Class-Balanced\_Loss\_Based\_on\_Effective\_Number\_of\_Samples\_CVPR\_2019\_paper.html}.

\bibitem[Deng et~al.(2009)Deng, Dong, Socher, Li, Li, and Fei{-}Fei]{DBLP:conf/cvpr/DengDSLL009}
Jia Deng, Wei Dong, Richard Socher, Li{-}Jia Li, Kai Li, and Li~Fei{-}Fei.
\newblock Imagenet: {A} large-scale hierarchical image database.
\newblock In \emph{2009 {IEEE} Computer Society Conference on Computer Vision and Pattern Recognition {(CVPR} 2009), 20-25 June 2009, Miami, Florida, {USA}}, pages 248--255. {IEEE} Computer Society, 2009.
\newblock \doi{10.1109/CVPR.2009.5206848}.
\newblock URL \url{https://doi.org/10.1109/CVPR.2009.5206848}.

\bibitem[He et~al.(2016)He, Zhang, Ren, and Sun]{DBLP:conf/cvpr/HeZRS16}
Kaiming He, Xiangyu Zhang, Shaoqing Ren, and Jian Sun.
\newblock Deep residual learning for image recognition.
\newblock In \emph{2016 {IEEE} Conference on Computer Vision and Pattern Recognition, {CVPR} 2016, Las Vegas, NV, USA, June 27-30, 2016}, pages 770--778. {IEEE} Computer Society, 2016.
\newblock \doi{10.1109/CVPR.2016.90}.
\newblock URL \url{https://doi.org/10.1109/CVPR.2016.90}.

\bibitem[Hong et~al.(2021)Hong, Han, Choi, Seo, Kim, and Chang]{DBLP:conf/cvpr/HongHCSKC21}
Youngkyu Hong, Seungju Han, Kwanghee Choi, Seokjun Seo, Beomsu Kim, and Buru Chang.
\newblock Disentangling label distribution for long-tailed visual recognition.
\newblock In \emph{{IEEE} Conference on Computer Vision and Pattern Recognition, {CVPR} 2021, virtual, June 19-25, 2021}, pages 6626--6636. Computer Vision Foundation / {IEEE}, 2021.
\newblock \doi{10.1109/CVPR46437.2021.00656}.
\newblock URL \url{https://openaccess.thecvf.com/content/CVPR2021/html/Hong\_Disentangling\_Label\_Distribution\_for\_Long-Tailed\_Visual\_Recognition\_CVPR\_2021\_paper.html}.

\bibitem[Horn et~al.(2018)Horn, Aodha, Song, Cui, Sun, Shepard, Adam, Perona, and Belongie]{DBLP:conf/cvpr/HornASCSSAPB18}
Grant~Van Horn, Oisin~Mac Aodha, Yang Song, Yin Cui, Chen Sun, Alexander Shepard, Hartwig Adam, Pietro Perona, and Serge~J. Belongie.
\newblock The inaturalist species classification and detection dataset.
\newblock In \emph{2018 {IEEE} Conference on Computer Vision and Pattern Recognition, {CVPR} 2018, Salt Lake City, UT, USA, June 18-22, 2018}, pages 8769--8778. Computer Vision Foundation / {IEEE} Computer Society, 2018.
\newblock \doi{10.1109/CVPR.2018.00914}.
\newblock URL \url{http://openaccess.thecvf.com/content\_cvpr\_2018/html/Van\_Horn\_The\_INaturalist\_Species\_CVPR\_2018\_paper.html}.

\bibitem[Jiang et~al.(2022)Jiang, Sun, Wang, and Liu]{fedlsr}
Xuefeng Jiang, Sheng Sun, Yuwei Wang, and Min Liu.
\newblock Towards federated learning against noisy labels via local self-regularization.
\newblock In Mohammad~Al Hasan and Li~Xiong, editors, \emph{Proceedings of the 31st {ACM} International Conference on Information {\&} Knowledge Management, Atlanta, GA, USA, October 17-21, 2022}, pages 862--873. {ACM}, 2022.
\newblock \doi{10.1145/3511808.3557475}.
\newblock URL \url{https://doi.org/10.1145/3511808.3557475}.

\bibitem[Jiang et~al.(2024{\natexlab{a}})Jiang, Li, Wu, Wu, Li, Sun, Xu, Wang, Li, and Liu]{fnbench}
Xuefeng Jiang, Jia Li, Nannan Wu, Zhiyuan Wu, Xujing Li, Sheng Sun, Gang Xu, Yuwei Wang, Qi~Li, and Min Liu.
\newblock Fnbench: Benchmarking robust federated learning against noisy labels.
\newblock 2024{\natexlab{a}}.
\newblock \doi{10.36227/techrxiv.172503083.36644691/v1}.
\newblock URL \url{http://dx.doi.org/10.36227/techrxiv.172503083.36644691/v1}.

\bibitem[Jiang et~al.(2024{\natexlab{b}})Jiang, Sun, Li, Xue, Li, Wu, Xu, Wang, and Liu]{fedelc}
Xuefeng Jiang, Sheng Sun, Jia Li, Jingjing Xue, Runhan Li, Zhiyuan Wu, Gang Xu, Yuwei Wang, and Min Liu.
\newblock Tackling noisy clients in federated learning with end-to-end label correction.
\newblock \emph{arXiv preprint arXiv:2408.04301}, 2024{\natexlab{b}}.

\bibitem[Jin et~al.(2022)Jin, Chen, Gu, and Li]{DBLP:conf/icpads/JinCGL22}
Cheng Jin, Xuandong Chen, Yi~Gu, and Qun Li.
\newblock Feddyn: {A} dynamic and efficient federated distillation approach on recommender system.
\newblock In \emph{28th {IEEE} International Conference on Parallel and Distributed Systems, {ICPADS} 2022, Nanjing, China, January 10-12, 2023}, pages 786--793. {IEEE}, 2022.
\newblock \doi{10.1109/ICPADS56603.2022.00107}.
\newblock URL \url{https://doi.org/10.1109/ICPADS56603.2022.00107}.

\bibitem[Kang et~al.(2020)Kang, Xie, Rohrbach, Yan, Gordo, Feng, and Kalantidis]{DBLP:conf/iclr/KangXRYGFK20}
Bingyi Kang, Saining Xie, Marcus Rohrbach, Zhicheng Yan, Albert Gordo, Jiashi Feng, and Yannis Kalantidis.
\newblock Decoupling representation and classifier for long-tailed recognition.
\newblock In \emph{8th International Conference on Learning Representations, {ICLR} 2020, Addis Ababa, Ethiopia, April 26-30, 2020}. OpenReview.net, 2020.
\newblock URL \url{https://openreview.net/forum?id=r1gRTCVFvB}.

\bibitem[Krizhevsky(2009)]{Krizhevsky2009LearningML}
Alex Krizhevsky.
\newblock Learning multiple layers of features from tiny images.
\newblock 2009.
\newblock URL \url{https://api.semanticscholar.org/CorpusID:18268744}.

\bibitem[Li et~al.(2020)Li, Sahu, Zaheer, Sanjabi, Talwalkar, and Smith]{DBLP:conf/mlsys/LiSZSTS20}
Tian Li, Anit~Kumar Sahu, Manzil Zaheer, Maziar Sanjabi, Ameet Talwalkar, and Virginia Smith.
\newblock Federated optimization in heterogeneous networks.
\newblock In Inderjit~S. Dhillon, Dimitris~S. Papailiopoulos, and Vivienne Sze, editors, \emph{Proceedings of Machine Learning and Systems 2020, MLSys 2020, Austin, TX, USA, March 2-4, 2020}. mlsys.org, 2020.
\newblock URL \url{https://proceedings.mlsys.org/paper\_files/paper/2020/hash/1f5fe83998a09396ebe6477d9475ba0c-Abstract.html}.

\bibitem[Li et~al.(2021)Li, Jiang, Zhang, Kamp, and Dou]{DBLP:conf/iclr/LiJZKD21}
Xiaoxiao Li, Meirui Jiang, Xiaofei Zhang, Michael Kamp, and Qi~Dou.
\newblock Fedbn: Federated learning on non-iid features via local batch normalization.
\newblock In \emph{9th International Conference on Learning Representations, {ICLR} 2021, Virtual Event, Austria, May 3-7, 2021}. OpenReview.net, 2021.
\newblock URL \url{https://openreview.net/forum?id=6YEQUn0QICG}.

\bibitem[Li and Zhan(2021)]{DBLP:conf/kdd/LiZ21}
Xin{-}Chun Li and De{-}Chuan Zhan.
\newblock Fedrs: Federated learning with restricted softmax for label distribution non-iid data.
\newblock In Feida Zhu, Beng~Chin Ooi, and Chunyan Miao, editors, \emph{{KDD} '21: The 27th {ACM} {SIGKDD} Conference on Knowledge Discovery and Data Mining, Virtual Event, Singapore, August 14-18, 2021}, pages 995--1005. {ACM}, 2021.
\newblock \doi{10.1145/3447548.3467254}.
\newblock URL \url{https://doi.org/10.1145/3447548.3467254}.

\bibitem[Li et~al.(2023{\natexlab{a}})Li, Liu, Sun, Wang, Jiang, and Jiang]{fedtrip}
Xujing Li, Min Liu, Sheng Sun, Yuwei Wang, Hui Jiang, and Xuefeng Jiang.
\newblock Fedtrip: A resource-efficient federated learning method with triplet regularization.
\newblock In \emph{2023 IEEE International Parallel and Distributed Processing Symposium (IPDPS)}, pages 809--819. IEEE, 2023{\natexlab{a}}.

\bibitem[Li et~al.(2023{\natexlab{b}})Li, Sun, Liu, Ren, Jiang, and He]{fedcrac}
Xujing Li, Sheng Sun, Min Liu, Ju~Ren, Xuefeng Jiang, and Tianliu He.
\newblock Federated classification tasks in long-tailed data environments via classifier representation adjustment and calibration.
\newblock \emph{Authorea Preprints}, 2023{\natexlab{b}}.

\bibitem[Lin et~al.(2014)Lin, Maire, Belongie, Hays, Perona, Ramanan, Doll{\'{a}}r, and Zitnick]{DBLP:conf/eccv/LinMBHPRDZ14}
Tsung{-}Yi Lin, Michael Maire, Serge~J. Belongie, James Hays, Pietro Perona, Deva Ramanan, Piotr Doll{\'{a}}r, and C.~Lawrence Zitnick.
\newblock Microsoft {COCO:} common objects in context.
\newblock In David~J. Fleet, Tom{\'{a}}s Pajdla, Bernt Schiele, and Tinne Tuytelaars, editors, \emph{Computer Vision - {ECCV} 2014 - 13th European Conference, Zurich, Switzerland, September 6-12, 2014, Proceedings, Part {V}}, volume 8693 of \emph{Lecture Notes in Computer Science}, pages 740--755. Springer, 2014.
\newblock \doi{10.1007/978-3-319-10602-1\_48}.
\newblock URL \url{https://doi.org/10.1007/978-3-319-10602-1\_48}.

\bibitem[Lin et~al.(2017)Lin, Goyal, Girshick, He, and Doll{\'{a}}r]{DBLP:conf/iccv/LinGGHD17}
Tsung{-}Yi Lin, Priya Goyal, Ross~B. Girshick, Kaiming He, and Piotr Doll{\'{a}}r.
\newblock Focal loss for dense object detection.
\newblock In \emph{{IEEE} International Conference on Computer Vision, {ICCV} 2017, Venice, Italy, October 22-29, 2017}, pages 2999--3007. {IEEE} Computer Society, 2017.
\newblock \doi{10.1109/ICCV.2017.324}.
\newblock URL \url{https://doi.org/10.1109/ICCV.2017.324}.

\bibitem[Luo et~al.(2021)Luo, Chen, Hu, Zhang, Liang, and Feng]{DBLP:conf/nips/LuoCHZLF21}
Mi~Luo, Fei Chen, Dapeng Hu, Yifan Zhang, Jian Liang, and Jiashi Feng.
\newblock No fear of heterogeneity: Classifier calibration for federated learning with non-iid data.
\newblock In Marc'Aurelio Ranzato, Alina Beygelzimer, Yann~N. Dauphin, Percy Liang, and Jennifer~Wortman Vaughan, editors, \emph{Advances in Neural Information Processing Systems 34: Annual Conference on Neural Information Processing Systems 2021, NeurIPS 2021, December 6-14, 2021, virtual}, pages 5972--5984, 2021.
\newblock URL \url{https://proceedings.neurips.cc/paper/2021/hash/2f2b265625d76a6704b08093c652fd79-Abstract.html}.

\bibitem[McMahan et~al.(2017)McMahan, Moore, Ramage, Hampson, and y~Arcas]{DBLP:conf/aistats/McMahanMRHA17}
Brendan McMahan, Eider Moore, Daniel Ramage, Seth Hampson, and Blaise~Ag{\"{u}}era y~Arcas.
\newblock Communication-efficient learning of deep networks from decentralized data.
\newblock In Aarti Singh and Xiaojin~(Jerry) Zhu, editors, \emph{Proceedings of the 20th International Conference on Artificial Intelligence and Statistics, {AISTATS} 2017, 20-22 April 2017, Fort Lauderdale, FL, {USA}}, volume~54 of \emph{Proceedings of Machine Learning Research}, pages 1273--1282. {PMLR}, 2017.
\newblock URL \url{http://proceedings.mlr.press/v54/mcmahan17a.html}.

\bibitem[Menon et~al.(2021)Menon, Jayasumana, Rawat, Jain, Veit, and Kumar]{DBLP:conf/iclr/MenonJRJVK21}
Aditya~Krishna Menon, Sadeep Jayasumana, Ankit~Singh Rawat, Himanshu Jain, Andreas Veit, and Sanjiv Kumar.
\newblock Long-tail learning via logit adjustment.
\newblock In \emph{9th International Conference on Learning Representations, {ICLR} 2021, Virtual Event, Austria, May 3-7, 2021}. OpenReview.net, 2021.
\newblock URL \url{https://openreview.net/forum?id=37nvvqkCo5}.

\bibitem[Mohassel and Zhang(2017)]{DBLP:conf/sp/MohasselZ17}
Payman Mohassel and Yupeng Zhang.
\newblock Secureml: {A} system for scalable privacy-preserving machine learning.
\newblock In \emph{2017 {IEEE} Symposium on Security and Privacy, {SP} 2017, San Jose, CA, USA, May 22-26, 2017}, pages 19--38. {IEEE} Computer Society, 2017.
\newblock \doi{10.1109/SP.2017.12}.
\newblock URL \url{https://doi.org/10.1109/SP.2017.12}.

\bibitem[Nazir and Kaleem(2023)]{diagnostics13091532}
Sajid Nazir and Mohammad Kaleem.
\newblock Federated learning for medical image analysis with deep neural networks.
\newblock \emph{Diagnostics}, 13\penalty0 (9), 2023.
\newblock ISSN 2075-4418.
\newblock \doi{10.3390/diagnostics13091532}.
\newblock URL \url{https://www.mdpi.com/2075-4418/13/9/1532}.

\bibitem[Nguyen et~al.(2022)Nguyen, Do, Tran, Nguyen, Duong, Phan, Tjiputra, and Tran]{DBLP:conf/ivs/NguyenDTNDPTT22}
Anh Nguyen, Tuong Do, Minh Tran, Binh~X. Nguyen, Chien Duong, Tu~Phan, Erman Tjiputra, and Quang~D. Tran.
\newblock Deep federated learning for autonomous driving.
\newblock In \emph{2022 {IEEE} Intelligent Vehicles Symposium, {IV} 2022, Aachen, Germany, June 4-9, 2022}, pages 1824--1830. {IEEE}, 2022.
\newblock \doi{10.1109/IV51971.2022.9827020}.
\newblock URL \url{https://doi.org/10.1109/IV51971.2022.9827020}.

\bibitem[Shang et~al.(2022{\natexlab{a}})Shang, Lu, Cheung, and Wang]{DBLP:conf/icmcs/ShangLCW22}
Xinyi Shang, Yang Lu, Yiu{-}Ming Cheung, and Hanzi Wang.
\newblock {FEDIC:} federated learning on non-iid and long-tailed data via calibrated distillation.
\newblock In \emph{{IEEE} International Conference on Multimedia and Expo, {ICME} 2022, Taipei, Taiwan, July 18-22, 2022}, pages 1--6. {IEEE}, 2022{\natexlab{a}}.
\newblock \doi{10.1109/ICME52920.2022.9860009}.
\newblock URL \url{https://doi.org/10.1109/ICME52920.2022.9860009}.

\bibitem[Shang et~al.(2022{\natexlab{b}})Shang, Lu, Huang, and Wang]{DBLP:conf/ijcai/ShangLHW22}
Xinyi Shang, Yang Lu, Gang Huang, and Hanzi Wang.
\newblock Federated learning on heterogeneous and long-tailed data via classifier re-training with federated features.
\newblock In Luc~De Raedt, editor, \emph{Proceedings of the Thirty-First International Joint Conference on Artificial Intelligence, {IJCAI} 2022, Vienna, Austria, 23-29 July 2022}, pages 2218--2224. ijcai.org, 2022{\natexlab{b}}.
\newblock \doi{10.24963/IJCAI.2022/308}.
\newblock URL \url{https://doi.org/10.24963/ijcai.2022/308}.

\bibitem[Wang et~al.(2020)Wang, Li, Kang, Li, Liew, Tang, Hoi, and Feng]{DBLP:conf/eccv/WangLKLLTHF20}
Tao Wang, Yu~Li, Bingyi Kang, Junnan Li, Jun~Hao Liew, Sheng Tang, Steven C.~H. Hoi, and Jiashi Feng.
\newblock The devil is in classification: {A} simple framework for long-tail instance segmentation.
\newblock In Andrea Vedaldi, Horst Bischof, Thomas Brox, and Jan{-}Michael Frahm, editors, \emph{Computer Vision - {ECCV} 2020 - 16th European Conference, Glasgow, UK, August 23-28, 2020, Proceedings, Part {XIV}}, volume 12359 of \emph{Lecture Notes in Computer Science}, pages 728--744. Springer, 2020.
\newblock \doi{10.1007/978-3-030-58568-6\_43}.
\newblock URL \url{https://doi.org/10.1007/978-3-030-58568-6\_43}.

\bibitem[Wu et~al.(2023)Wu, Yu, Jiang, Cheng, and Yan]{fednoro}
Nannan Wu, Li~Yu, Xuefeng Jiang, Kwang{-}Ting Cheng, and Zengqiang Yan.
\newblock Fednoro: Towards noise-robust federated learning by addressing class imbalance and label noise heterogeneity.
\newblock In \emph{Proceedings of the Thirty-Second International Joint Conference on Artificial Intelligence, {IJCAI} 2023, 19th-25th August 2023, Macao, SAR, China}, pages 4424--4432. ijcai.org, 2023.
\newblock \doi{10.24963/ijcai.2023/492}.
\newblock URL \url{https://doi.org/10.24963/ijcai.2023/492}.

\bibitem[Xiao et~al.(2024)Xiao, Chen, Liu, Feng, Wu, Liu, Zhou, Yang, and Liu]{DBLP:journals/corr/abs-2401-08977}
Zikai Xiao, Zihan Chen, Liyinglan Liu, Yang Feng, Jian Wu, Wanlu Liu, Joey~Tianyi Zhou, Howard~Hao Yang, and Zuozhu Liu.
\newblock Fedloge: Joint local and generic federated learning under long-tailed data.
\newblock \emph{CoRR}, abs/2401.08977, 2024.
\newblock \doi{10.48550/ARXIV.2401.08977}.
\newblock URL \url{https://doi.org/10.48550/arXiv.2401.08977}.

\bibitem[Zang et~al.(2021)Zang, Huang, and Loy]{DBLP:conf/iccv/ZangHL21}
Yuhang Zang, Chen Huang, and Chen~Change Loy.
\newblock {FASA:} feature augmentation and sampling adaptation for long-tailed instance segmentation.
\newblock In \emph{2021 {IEEE/CVF} International Conference on Computer Vision, {ICCV} 2021, Montreal, QC, Canada, October 10-17, 2021}, pages 3437--3446. {IEEE}, 2021.
\newblock \doi{10.1109/ICCV48922.2021.00344}.
\newblock URL \url{https://doi.org/10.1109/ICCV48922.2021.00344}.

\bibitem[Zeng et~al.(2022)Zeng, Semiari, Chen, Saad, and Bennis]{DBLP:journals/twc/ZengSCSB22}
Tengchan Zeng, Omid Semiari, Mingzhe Chen, Walid Saad, and Mehdi Bennis.
\newblock Federated learning on the road autonomous controller design for connected and autonomous vehicles.
\newblock \emph{{IEEE} Trans. Wirel. Commun.}, 21\penalty0 (12):\penalty0 10407--10423, 2022.
\newblock \doi{10.1109/TWC.2022.3183996}.
\newblock URL \url{https://doi.org/10.1109/TWC.2022.3183996}.

\bibitem[Zhang et~al.(2024)Zhang, Liu, Hua, and Cao]{DBLP:conf/aaai/ZhangLHC24}
Jianqing Zhang, Yang Liu, Yang Hua, and Jian Cao.
\newblock Fedtgp: Trainable global prototypes with adaptive-margin-enhanced contrastive learning for data and model heterogeneity in federated learning.
\newblock In Michael~J. Wooldridge, Jennifer~G. Dy, and Sriraam Natarajan, editors, \emph{Thirty-Eighth {AAAI} Conference on Artificial Intelligence, {AAAI} 2024, Thirty-Sixth Conference on Innovative Applications of Artificial Intelligence, {IAAI} 2024, Fourteenth Symposium on Educational Advances in Artificial Intelligence, {EAAI} 2014, February 20-27, 2024, Vancouver, Canada}, pages 16768--16776. {AAAI} Press, 2024.
\newblock \doi{10.1609/AAAI.V38I15.29617}.
\newblock URL \url{https://doi.org/10.1609/aaai.v38i15.29617}.

\bibitem[Zhang et~al.(2023)Zhang, Kang, Hooi, Yan, and Feng]{DBLP:journals/pami/ZhangKHYF23}
Yifan Zhang, Bingyi Kang, Bryan Hooi, Shuicheng Yan, and Jiashi Feng.
\newblock Deep long-tailed learning: {A} survey.
\newblock \emph{{IEEE} Trans. Pattern Anal. Mach. Intell.}, 45\penalty0 (9):\penalty0 10795--10816, 2023.
\newblock \doi{10.1109/TPAMI.2023.3268118}.
\newblock URL \url{https://doi.org/10.1109/TPAMI.2023.3268118}.

\bibitem[Zhang and Pfister(2021)]{DBLP:conf/iccv/ZhangP21}
Zizhao Zhang and Tomas Pfister.
\newblock Learning fast sample re-weighting without reward data.
\newblock In \emph{2021 {IEEE/CVF} International Conference on Computer Vision, {ICCV} 2021, Montreal, QC, Canada, October 10-17, 2021}, pages 705--714. {IEEE}, 2021.
\newblock \doi{10.1109/ICCV48922.2021.00076}.
\newblock URL \url{https://doi.org/10.1109/ICCV48922.2021.00076}.

\end{thebibliography}

\end{document}